\documentclass[10pt, a4paper]{article}

\usepackage[]{lrec-coling2024} 

\usepackage[utf8]{inputenc}
\pdfoutput=1
\usepackage{times}
\usepackage{latexsym}
\usepackage{amsmath}
\usepackage{amssymb}
\usepackage{graphicx}
\usepackage{tabularx}
\usepackage{multirow}
\usepackage{booktabs}
\usepackage{enumitem}
\usepackage{natbib}
\usepackage{makecell}
\usepackage{url}

\title{CTSM: Combining Trait and State Emotions for Empathetic Response Model}

\name{Yufeng Wang$^{1,2}$, Chao Chen$^{3}$, Zhou Yang$^{1,2}$, Shuhui Wang$^{1,2}$, Xiangwen Liao$^{1,2{\ast}}$\thanks{*Corresponding author.}} 

\address{
$^1$College of Computer and Data Science, Fuzhou University, Fuzhou, China\\
$^2$Digital Fujian Institute of Financial Big Data, Fuzhou, China\\
$^3$School of Computer Science and Technology, Harbin Institute of Technology (Shenzhen), China\\
         211027083@fzu.edu.cn, cha01nbox@gmail.com\\
         \{200310007, 221027214, liaoxw\}@fzu.edu.cn
}

\abstract{
Empathetic response generation endeavors to empower dialogue systems to perceive speakers' emotions and generate empathetic responses accordingly. Psychological research demonstrates that emotion, as an essential factor in empathy, encompasses trait emotions, which are static and context-independent, and state emotions, which are dynamic and context-dependent.  However, previous studies treat them in isolation, leading to insufficient emotional perception of the context, and subsequently, less effective empathetic expression. To address this problem, we propose \textbf{C}ombining \textbf{T}rait and \textbf{S}tate emotions for Empathetic Response \textbf{M}odel (\textbf{CTSM}). Specifically, to sufficiently perceive emotions in dialogue, we first construct and encode trait and state emotion embeddings, and then we further enhance emotional perception capability through an emotion guidance module that guides emotion representation. In addition, we propose a cross-contrastive learning decoder to enhance the model's empathetic expression capability by aligning trait and state emotions between generated responses and contexts. Both automatic and manual evaluation results demonstrate that CTSM outperforms state-of-the-art baselines and can generate more empathetic responses. Our code is available at \url{https://github.com/wangyufeng-empty/CTSM}
 \\ \newline \Keywords{
empathetic response generation, dialogue system, emotion recognition, contrastive learning
 }}

\begin{document}
\maketitleabstract

\section{Introduction}

\begin{figure}[!t]
\begin{center}
\includegraphics[width=\columnwidth]{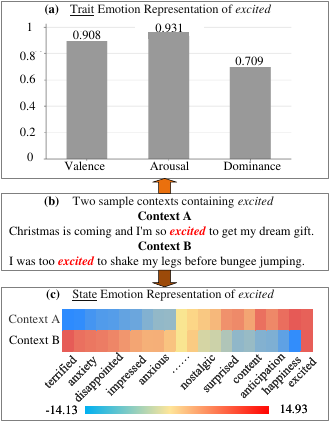} 
\caption{An example of trait and state emotions. (a) The bar chart illustrates the context-independent trait emotions of \textit{excited}. (b) Two sample contexts contain the term \textit{excited}. (c) The heat map displays the varying state emotions of \textit{excited} across two different contexts, with warmer colors indicating stronger emotion inclination and cooler colors denoting lesser emotional intensity.}
\label{fig.1}
\end{center}
\end{figure}

Empathy is crucial in human conversation \citep{1.1} and human-like dialogue systems \citep{1.3-9961311, 1.2-zhao-etal-2023-dont}. Central to this work is the empathetic response generation task \citep{2-ED}, which aims to produce empathetic responses by profoundly comprehending speakers' emotions \citep{1.2-zhao-etal-2023-dont}. Emotion, as an essential factor facilitating empathy \citep{1.6-情绪和共情关系pmid25706828, 1.4-情绪和共情关系doi:10.1146/annurev-psych-010419-050830,1.5-情绪和共情关系PMID:33570970}, bridges communicators and fosters understanding \citep{4.1}. Psychological studies \citep{5} differentiate between trait (static and context-independent \citep{6}) and state (dynamic and context-dependent \citep{6,7}) emotions. Specifically, we regard \textit{static} as the inherent emotional connotation of textual words, whereas \textit{dynamic} corresponds to the emotion's variability and adaptability to contexts.
Figure \ref{fig.1} illustrates the distinction between trait and state emotions. 
The bar chart highlights that the trait emotion of \textit{excited} consistently embodies the fundamental emotional dimensions of Valence, Arousal, and Dominance \cite{32-mohammad-2018-obtaining} in context A and B. 
Trait emotions can be accurately quantified and remain context-independent, and overlooking them may miss inherent emotional connotations of words, weakening emotion understanding. Conversely, the heat map presents the diverse emotional expressions of \textit{excited} across various contexts. In context A, it conveys positive feelings like \textit{happiness} and \textit{anticipation}, while in context B, \textit{excited} tends towards negative emotions like \textit{terrified} and \textit{anxiety}. Ignoring state emotions could confuse semantics and emotional interpretation.

Existing approaches focus separately on perceiving only one type of emotion. Research targeting trait emotions often utilizes pre-trained classifiers \citep{2-ED} and external knowledge \citep{9-KEMP,10-CEM,2023-case,2023-EMPSOA}. Conversely, approaches centered on state emotions employ multi-listener frameworks \citep{11-MOEL}, emotion mimicry techniques \citep{12-MIME-majumder-etal-2020-mime}, and embedding adjustments \citep{29-agrawal-etal-2018-learning,30-mao2019sentiment,31-10.1145/3269206.3269284}. However, treating trait and state emotions in isolation comprised the completeness and intricacy of emotional expression in dialogue. Neglecting the emotional reaction \citep{emotional-reaction} stemming from the interaction between trait and state emotions can engender inaccurate emotion comprehension and categorization, generating inappropriate empathetic responses. Therefore, modeling both emotion types is imperative for empathetic response generation, but remains underexplored.

To this end, we propose a \textbf{C}ombining \textbf{T}rait and \textbf{S}tate Emotions for Empathetic Response \textbf{M}odel (CTSM) to fully incorporate trait and state emotions, enabling more comprehensive perception and expression of contextual emotions. 
First, in addressing the distinction between static trait emotions and dynamic state emotions, we present two specialized embedding patterns to capture their unique characteristics at the token level. Building on this foundation, we introduce emotion encoders as the essential component to extract and refine the nuanced feature representations inherent in these embeddings.
Subsequently, to enhance emotion perception capability, we propose an emotion guidance module with teacher and student components. The teacher guides the student through emotion labels to enhance the student's understanding of complex emotions.
Additionally, we design a cross-contrastive learning decoder to enhance CTSM's empathetic expression capability that aligns the features of generated responses and contexts in terms of trait and state emotions. 

Experiments with benchmark models on the E{\small MPATHETIC}D{\small IALOGUES} (ED) dataset \citeplanguageresource{lr-ED} demonstrate that CTSM outperforms benchmark models, particularly excelling in emotion accuracy and diversity metrics. Further studies verify that CTSM can accurately perceive both trait and state emotions.

Our contributions are summarized as follows:

\begin{itemize}[leftmargin=*]
    \item{To the best of our knowledge, our work is the first to simultaneously model both trait and state emotions for \textit{each} token within the dialogue text. This addresses the limitations in emotion perception methodologies of prior works.
}
    \item{We augment the interplay between trait and state emotions utilizing an emotion guidance module, which improves the perception of intricate emotions through full emotion feature guidance. Furthermore, we employ a cross-contrastive learning decoder to enhance empathetic expression during empathetic response generation.
}
    \item {The experimental results demonstrate that CTSM effectively combines trait and state emotions within dialogues, exhibiting enhanced empathetic capabilities.}

\end{itemize}

\section{Related Work}

Empathetic response generation involves perceiving emotions conveyed by speakers in a dialogue to produce sympathetic responses \cite{2-ED}. Early approaches directly simulated emotions using rules and statistics \citep{20-COLBY1972199,21-10.1007/978-3-642-24571-8_2,19-ADAMOPOULOU2020100006}, but faced challenges in scalability, flexibility, and cost. 
Recent methods, leveraging the power of deep neural networks \citep{14DBLP:journals/corr/abs-1301-3781,15pennington-etal-2014-glove,13MA202050} and word embeddings \cite{26-8244338,24-baali2019emotion,25-10.1007/978-3-319-99010-1_31}, have made significant strides in perceiving emotions in conversational text. Building on these advancements, state-of-the-art approaches can be broadly categorized into two groups: those targeting the perception of trait emotions and those focusing on state emotions within dialogue contexts.

The first category of methods emphasizes the perception of trait emotions to enhance emotional accuracy. Specifically, \citet{2-ED} leveraged a pre-trained emotion classifier to capture trait emotions in context, and \citet{8-EMPDG} adopted interactive discriminators to extract multi-resolution trait emotions. Further considering specific words like negation combined with word intensity \citep{27-Zhong_Wang_Miao_2019} and words causing emotional causality \citep{28-kim-etal-2021-perspective} can strengthen the model's perception of subtle trait emotions. However, the lack of external knowledge makes it challenging for models to perceive implicit emotions. \citet{9-KEMP} addressed this limitation by utilizing external knowledge to construct an emotion context graph, enhancing the expression of implicit trait emotions in the semantic space. \citet{10-CEM} built COMET through external reasoning knowledge, reinforcing the perception of trait emotions. Building on this, \citet{2023-case} and \citet{2023-EMPSOA} integrate external knowledge through graph structure to enhance the model's empathetic capabilities. However, these models focus on perceiving static, context-independent trait emotions while neglecting dynamic state emotions, leading to a misalignment between contextual semantics and the emotions conveyed in the text.

The second type of method focuses on perceiving state emotions to enrich emotion understanding, such as modeling mixed emotions using multiple listeners \citep{11-MOEL} or mimicking user emotions considering emotion polarity and randomness \citep{12-MIME-majumder-etal-2020-mime}. However, directly modeling global context can cause semantically similar words to convey opposing emotions \citep{29-agrawal-etal-2018-learning}. Thus, some methods address this by constructing contextual word embeddings that capture emotional influences on individual words \citep{29-agrawal-etal-2018-learning,30-mao2019sentiment,31-10.1145/3269206.3269284,2023-escm}, enabling richer state emotion perception. However, these approaches overlook the inherent static trait emotions within the dialogue, leading to inaccurate discernment of contextual emotions and generating inappropriate empathetic responses.


\section{Method}
\begin{figure*}[t]
\centering
\includegraphics[width=\dimexpr\textwidth]{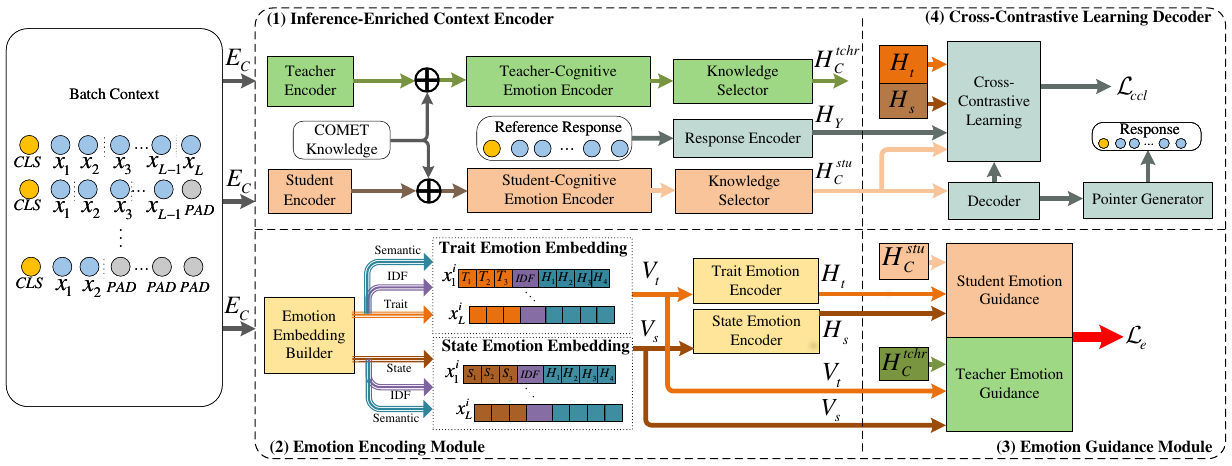}
\caption{\label{fig CTSM}
An overview architecture of CTSM. It consists of four parts: 
\textbf{1)} Inference-Enriched Context Encoder; \textbf{2)} Emotion Encoding Module; \textbf{3)} Emotion Guidance Module; \textbf{4) }Cross-Contrastive Learning Decoder.}
\end{figure*}
\subsection{Overview}

Figure \ref{fig CTSM} illustrates the overall architecture of CTSM. To effectively perceive and utilize the trait and state emotions in dialogues, we abstract the model into four primary components: \textbf{1)} \textbf{Inference-Enriched Context Encoder} encodes the context and integrates inference knowledge; \textbf{2) Emotion Encoding Module} constructs and encodes two emotion embeddings, enabling the model to perceive both trait and state emotions from context fully; \textbf{3) Emotion Guidance Module} facilitates CTSM's learning of emotion representations by utilizing the inference-enriched context to enhance its emotion perception capabilities; \textbf{4)} \textbf{Cross-Contrastive Learning Decoder} employs cross-contrastive learning after decoding process during training, allowing CTSM to generate more empathetic and appropriate responses. 

\subsection{Task Formulation}

Given a dialogue history $D = [u_1, u_2, \dots, u_n]$ with a context-level emotion label \(\varepsilon\), as well as a set of emotion words \(e\), our goal is to generate empathetic responses $R=\left[r_1, r_2, \cdots, r_m\right]$ whose semantics and emotions align with the context while conveying empathy. \(e\) is the union of all emotion labels and $\varepsilon \in e$. \(D\) is made up of \(n\) sentences, and \(R\) contains \(m\) words. The $i$-th sentence $u_i = \left[x_1^i,x_2^i,\cdots,x_{l_i}^i\right]$ consists of $l_i$ words. 
For batches, sequences shorter than the maximum length $L$ are padded to $L$ with $[PAD]$ tokens.

\subsection{Inference-Enriched Context Encoder \label{sec:i-e-encoder}}

Following prior works \cite{9-KEMP,10-CEM}, we flatten the dialogue context, and prepend a $[CLS]$ token to obtain the context sequence $C=[C L S] \oplus u_1 \oplus u_2 \oplus \cdots \oplus u_{n}$, where $\oplus$ represents concatenation. The context embedding $E_C \in \mathbb{R}^{L\times d}$ is the sum of the word embeddings, positional embeddings, and dialogue state embeddings. An encoder is used to extract the contextual hidden representation from $E_C$:
\begin{equation}
H_C=\boldsymbol{\operatorname{Encoder}}_C\left(E_C\right),
\end{equation}
where $H_C \in \mathbb{R}^{L \times d}$ and $d$ is the dimension of the context encoding.

Referring to the approach of \citet{10-CEM} for the fusion of context and inferential knowledge, we establish the inference-enriched context teacher $H_C^{tchr}$ and similarly the student $H_C^{stu}$. 
These two contexts will be used in the emotion guidance module (in Sec.  \ref{Emotion Self-Distillation Guidance Module}) and cross-contrastive learning decoder (in Sec.  \ref{cross-contrastive learning based decoder}).

\subsection{Emotion Encoding Module}

In this subsection, we illustrate how to perceive trait and state emotions by considering their unique characteristics.  We also present how we integrate and encode them with external knowledge and the importance of words.

\subsubsection{Trait Emotions Encoding}

To ensure that the trait emotion embedding $V_t$ effectively captures the context-independent emotion, importance of words, and contextual semantics.
$V_t$ integrates static emotion knowledge from the VAD emotion lexicon \citeplanguageresource{lr-vad} $V_{VAD}$,
Inverse Document Frequency (IDF) \citep{IDF-10.5555/106765.106782} $V_{IDF}$ and condensed contextual semantics $\widetilde{H}_{C}$.
Formally,

\begin{equation}
V_t=V_{V A D} \oplus V_{IDF} \oplus \widetilde{H}_{C},
\end{equation}
\begin{equation}
\widetilde{H}_{C}=W_{C} H_C,
\end{equation}
where $V_{V A D} \in \mathbb{R}^{L \times 3}, V_{IDF} \in \mathbb{R}^{L \times 1}, \tilde{H}_C \in \mathbb{R}^{L \times d_{cs}}$ and $d_{c s}$ is the dimension after semantic compression. 
The VAD lexicon delineates emotions into three dimensions: Valence (negativity or positivity), Arousal (calmness or excitement), and Dominance (weak or strong control), with each value ranging from [0,1]. 
The words excluded in VAD are set with neutral default values [0.00, 0.50, 0.00].
Then, $V_t\in \mathbb{R}^{L \times d_t}$ is encoded to encapsulate the trait emotions representation within the context:
\begin{equation}
H_{t}=\boldsymbol{\operatorname{ Encoder}}_{t}\left(V_t\right),
\end{equation}
where $\boldsymbol{\operatorname{ Encoder}}_{t}\left(\cdot\right)$ is the trait emotion encoder and $H_{t} \in \mathbb{R}^{L \times d_t}$.  

\subsubsection{State Emotion Encoding \label{state emotion encoding}}

Regarding the state emotion embedding $V_s$, we initially define the dynamic state inclination $V_{\cos}$, whose entries with higher values indicate a stronger emotional inclination.
Specifically, \(V_{\cos}\) is the cosine similarity between the linear embeddings of emotion words \(\tilde{E}_{e}\) and context \(\tilde{E}_C\).

\begin{equation}
\widetilde{E}_e=W_1 \times \boldsymbol{\operatorname{ Embedding }} (e)+b_1,
\end{equation}
\begin{equation}
\widetilde{E}_C=W_2  E_C+b_2,
\end{equation}
\begin{equation}
V_{\cos }=\boldsymbol{\operatorname{\cos }}\left(\widetilde{E}_C, {\widetilde{E}_e}^\top\right),
\end{equation}
where $\widetilde{E}_e \in \mathbb{R}^{32 \times d}$, $\widetilde{E}_C \in \mathbb{R}^{L\times d}$ and $V_{cos} \in \mathbb{R}^{L \times 32}$.
The 32 dimensions in $\widetilde{E}_{e}$ correspond to the 32 emotion categories in the ED dataset, used for prediction classification. Notably, these emotion labels are \textit{not} fed into the emotion encoder.
$\boldsymbol{\operatorname{\cos }}(\cdot)$ is the cosine similarity function. $W_1, W_2, b_{{1}}$, $b_2$ are all trainable parameters.  
Next, $V_s$ consolidates the state inclination, IDF vector, and semantics \(\tilde{H}_C\) to understand the token's state emotions comprehensively.
\begin{equation}
V_s=V_{\cos} \oplus V_{IDF} \oplus \widetilde{H}_{C},
\end{equation}
where $V_s \in \mathbb{R}^{L \times d_s}$, $d_s$ is the dimensionality of the state emotion embedding. Finally, we encode $V_s$ to capture the state emotion representation:
\begin{equation}
H_{s}=\boldsymbol{\operatorname { Encoder }}_{s}\left(V_s\right),
\end{equation}
where $\boldsymbol{\operatorname{ Encoder}}_{s}\left(\cdot\right)$ is the state emotion encoder and $H_{s} \in \mathbb{R}^{L \times d_s}$.

\subsection{Emotion Guidance Module \label{Emotion Self-Distillation Guidance Module}}

In light of the intricacies of textual emotions in dialogues, a hard label may not encompass the blend of trait and state emotions. Drawing from research \citep{distill-searchC,distill-searchA,distill-searchB}, we design an emotion guidance module that enhances the model's capacity to perceive and comprehend intricate emotions. Specifically, we employ a teacher component to extract soft labels encapsulating the \textit{dark knowledge} \citep{dark-knowledge-44873}, which comprises hidden knowledge crossing both trait and state emotions.
The student model assimilates the dark knowledge by training with soft labels, leading to enhanced generalization and performance \citep{function-hahn-choi-2019-self}. Armed with augmented capabilities, the student is employed in both the emotion prediction (in Sec.  \ref{Student Emotion Guidance}) and decoding phases (in Sec.  \ref{cross-contrastive learning based decoder}).

\subsubsection{Teacher Emotion Guidance \label{Teacher Emotion Guidance}}
After getting the embedding of trait and state emotions as well as the teacher inference-enriched context, we concatenate them to $V_{t c h r}$. Subsequently, the teacher's semantic-emotion context \( C_{tchr} \) is derived by weighting \( V_{tchr} \) with the emotion intensities \( I \) \citep{9-KEMP}.
\begin{equation}
V_{t c h r}=H_C^{tchr} \oplus V_t \oplus V_s,
\end{equation}
\begin{equation}
C_{t c h r}=\boldsymbol{\operatorname{SUM}}_{d=1}\left(\boldsymbol{\sigma}\left(I\right) \times V_{tchr}\right),
\end{equation}
where $V_{tchr} \in \mathbb{R}^{L \times d_s}$, $C_{tchr}\in\ \mathbb{R}^{L\ \times\ d_s\ }$, $H_C^{tchr} \in \mathbb{R}^{L \times d}$. $\boldsymbol{\operatorname{{SUM}}}_{d=1}\left(\cdot\right)$ represents a summation over the first dimension to aggregate contextual semantics representations. $\boldsymbol{\sigma}\left(\cdot\right)$ is the softmax function. 
The emotion distribution $P_e^{tchr}$ predicted by the teacher, which captures the characteristics of both trait and state emotions, is given by:
\begin{equation}
\label{eq:s}
S=\boldsymbol{\sigma}\left(W_3^s\left(\boldsymbol{\operatorname {Tanh}}\left(W_3^c C_{tchr}+b_3\right)\right)\right),
\end{equation}
\begin{equation}
S'=\boldsymbol{\operatorname{Tanh}}\left({W}_{4}\left(C_{tchr}S\right)+{b}_{{4}}\right),
\end{equation}
\begin{equation}
\label{eq:p_e^tchr}
P_{e}^{tchr}=\boldsymbol{\sigma}\left(W_{ out}S'+ b_{out}\right),
\end{equation}
where $S \in \mathbb{R}^{L \times d}$, $S' \in \mathbb{R}^{L \times {d_s}}$, $P_{e}^{tchr} \in \mathbb{R}^{32}$. $W_3^s$, $W_3^c$, $ W_4, b_3, b_4$, $b_{out }$, and $W_{out}$ are all trainable parameters. The teacher's parameters are optimized by the Cross-Entropy Loss between teacher emotion prediction $P_{e}^{tchr}$ and the ground truth label $e^\ast$.
\begin{equation}
    \mathcal{L}_{t c h r}=-\log \left(P_{e}^{tchr}\left(e^\ast\right)\right).
\end{equation}

\subsubsection{Student Emotion Guidance \label{Student Emotion Guidance}}

The student shares a similar model structure with the teacher. However, the student concatenates the student inference-enriched context $H_C^{stu}$ with emotion representations $H_t$ and $H_s$, rather than $V_t$ and $V_s$. Specifically:
\begin{equation}
V_{stu}=H_C^{stu} \oplus H_t \oplus H_s,
\end{equation}
\begin{equation}
C_{stu}=\boldsymbol{\operatorname{SUM}}_{\boldsymbol{d}=\mathbf{1}}\left(\boldsymbol{\sigma}\left(I\right) \times V_{stu}\right),
\end{equation}
where $V_{stu} \in \mathbb{R}^{L \times d_s}$. $C_{stu} \in \mathbb{R}^{L \times d_s}$ is the student semantic-emotion context. 
Analogous to how $P_e^{tchr}$ is computed by Eqs. (\ref{eq:s}) - (\ref{eq:p_e^tchr}),
we obtain the student's emotion prediction $P_e^{stu}\in\mathbb{R}^{32}$ and employ it for predicting dialogue emotion, formally represented by the equation: $\hat{e}=\boldsymbol{\operatorname{argmax}}\left(P_{e}^{s t u}\right)$.  
To learn hidden knowledge from the teacher, the student is trained with soft labels:
\begin{equation}
\mathcal{L}_{s t u}=-\log \left(P_{e}^{stu}\left(P_{e}^{tchr}\right)\right).
\end{equation}
Ultimately, the objective of the teacher and student components concerning the emotion perception is:
\begin{equation}
\mathcal{L}_{e}=\mathcal{L}_{t c h r}+\mathcal{L}_{s t u}.
\end{equation}

\subsection{Cross-Contrastive Learning Decoder \label{cross-contrastive learning based decoder}}

Contrastive learning \cite{17pmlr-v119-chen20j,18-Sun_Shi_Gao_Ren_de_Rijke_Ren_2023} minimizes distances between positive samples while maximizing distances between negative samples. Inspired by its application in feature alignment \citep{35-10.1145/3488560.3498514}, we incorporate cross-contrastive learning into the decoding process. 
Specifically, we align dialogue emotions between responses and contexts by minimizing the distance among generated responses, target responses, contextual semantics, as well as trait and state emotions. 
Consequently, this enhances the student's contextual semantic representation and teacher-guided performance, further improving the model's ability for empathetic expression.

\subsubsection{Response Generation}

As mentioned in Sec. \ref{sec:i-e-encoder}, the student inference-enriched context $H_C^{stu}$ is used to make prediction for word distribution $P_w$:

\begin{equation}
\begin{aligned}
    P_w & = P\left(R_j \mid E_{R<j}, C,H_{t}, H_{s}\right) \\ 
    & =\boldsymbol{\operatorname{PoGen}}\left(\boldsymbol{\operatorname {Decoder}}\left(E_Y, H_C^{stu}\right)\right),
\end{aligned}
\end{equation}
where $E_{R<j}$ is the embedding of the generated responses up to time step $j-1$. $E_Y=\boldsymbol{\operatorname { Embedding }}(Y)$ is the embedding of the target response $Y$, and \(\boldsymbol{\operatorname{PoGen}}\left(\cdot\right)\) signifies the pointer generator network module \citep{36-see-etal-2017-get}.

Ultimately, the generation loss of the model is defined by a standard negative log-likelihood:
\begin{equation}
\mathcal{L}_{{g}}=-\sum_{j=1}^T \log P\left(R_j \mid E_{R<j}, C,H_{t}, H_{s}\right),
\end{equation}

\subsubsection{Cross-Contrastive Learning \label{Cross-Contrastive Learning}}

We adopt Contrastive Learning to align dialogue emotion representations between responses and contexts for the same context within a batch.
Besides aligning representations of contextual semantics $H_C^{stu}$ and generated responses $P_w$,
we are also interested in the hidden representation of target response $H_Y = \boldsymbol{\operatorname { Encoder }}_Y(E_Y)$, and the combined representation of trait and state emotions \( H_{ts} = H_{t} \oplus H_{s} \).
Then, by \textit{crossly} pairing these representations with each other, 
the set of positive sample pairs denoted as $\mathcal{H}^{+}$, consists of the five sample pairs for each context.
To be specific:
\begin{equation}
    \begin{aligned}
        \mathcal{H}^+ =\{&(H_Y,H_{ts}), (H_C^{stu},P_w), (H_C^{stu},H_Y),\\ &(H_{ts},P_w), (H_Y,P_w)\}, 
    \end{aligned}
\end{equation}
Notably, considering the tight correlations between emotions and semantics, our model not only aligns them within the current context but also captures the emotion correlations across various contexts in the batch. To avoid potential misalignment of emotions with semantics from other contexts, we exclude the pair \( (H_C^{stu}, H_{ts}) \).
Conversely, the representations of the different contexts are regarded as negative pairs, and the set of negative sample pairs $\mathcal{H}^-$ contains all such negative pairs.

The training objective for any given positive sample pair is to minimize the distance between their representations while maximizing the distance for negative pairs, expressed as:
\begin{equation}
\mathcal{L}_{cl}(h_p,h_q) = -\log \frac{e^{\boldsymbol{\operatorname{sim}}\left(h_p, h_q\right) / \tau}}{\sum_{\left(h_p,h_k\right)} e^{\boldsymbol{\operatorname{sim}}\left(h_p, h_k\right) / \tau}},
\end{equation}
where $\left(h_p, h_q\right) \in \mathcal{H}^+$, $\left(h_p,h_k\right) \in \mathcal{H}^-$. \( \boldsymbol{\operatorname{sim}}(\cdot) \) computes similarity using the dot product. \( \tau \) is a temperature parameter adjusting the scale of similarity scores. The loss \( \mathcal{L}_{ccl} \) is then computed as the average loss across the five positive sample pairs:
\begin{equation}
\mathcal{L}_{ccl} =\frac{1}{5}\sum_{\left(h_p,h_q\right) } \mathcal{L}_{cl}(h_p,h_q).
\end{equation}
Integrating the diversity loss \({\mathcal{L}}_{div}\) suggested by \citet{10-CEM}, the total loss of our model is the weighted sum of the four mentioned losses:
\begin{equation}
\begin{aligned}
\mathcal{L}  =\gamma_1\mathcal{L}_{e}+\gamma_2\mathcal{L}_{{g}} +\gamma_3 \mathcal{L}_{{ccl}}+\gamma_4\mathcal{L}_{{div }},
\end{aligned}
\end{equation}
where $\gamma_1$, $\gamma_2$, $\gamma_3$, and $\gamma_4$ are hyperparameters that can be manually set.

\section{Experimental Settings}

\subsection{Baselines for Comparison}

We compare the proposed model with six state-of-the-art (SOTA) benchmark models:

\begin{itemize}[left=0pt,itemsep=0pt, parsep=0pt]
    \item \textbf{Transformer }\cite{34-10.5555/3295222.3295349}: is a vanilla Transformer-based model with encoder-decoder architecture for generation.
    \item \textbf{MoEL} \cite{11-MOEL}: is a Transformer-based empathetic response generation model using separate emotion decoders and a global contextual decoder to combine emotions softly.
    \item \textbf{MIME} \cite{12-MIME-majumder-etal-2020-mime}: is a Transformer-based model that considers emotion clustering based on polarity and emotion mimicry to generate empathetic responses.
    \item \textbf{EmpDG }\cite{8-EMPDG}: combines dialog-level and token-level emotions through a multi-resolution adversarial model with multi-granularity emotion modeling and user feedback.
    \item \textbf{KEMP} \cite{9-KEMP}: uses ConceptNet \citep{39-10.5555/3298023.3298212} to construct an emotion context graph, capturing implicit emotions to enrich representations for appropriate response generation.
    \item \textbf{CEM} \cite{10-CEM}: incorporates affection and cognition, and uses reasoning knowledge about the user's situation to enhance its ability to perceive and express emotions.
    \item \textbf{CASE} \citep{2023-case}: introduces commonsense reasoning and emotional concepts, aligning with the user's cognition and emotions at both coarse-grained and fine-grained levels to generate empathetic responses rich in information.
\end{itemize}

\subsection{Implementation Details}

We conduct experiments on E{\small MPATHETIC}D{\small IALOG\-UES} dataset, using the 8:1:1 train/validation/test split as in \citeplanguageresource{lr-ED}. CTSM uses 300-dimensional pre-trained GloVe vectors \citep{15pennington-etal-2014-glove} for embedding. The dynamic state inclination is 32-dimensional, VAD vectors are 3-dimensional, and the compressed semantic dimensions $d_{c s}$ are 10-dimensional. We set the cross-contrastive learning temperature $\tau$ to 0.07 and loss weights $\gamma_1$ to $\gamma_4$ as 1, 1, 1, and 1.5, respectively. When trained on a Tesla T4 GPU with a batch size of 16, our model utilizes the Adam optimizer \cite{41-kinga2015method} combined with the NoamOpt \cite{34-10.5555/3295222.3295349} learning rate schedule. The model converges after roughly 17,250 iterations.

\subsection{Evaluation Metrics}

\subsubsection{Automatic Evaluations}

We adopt four automated metrics for evaluation: Emotion Accuracy (\textbf{Acc}), Perplexity (\textbf{PPL}) \cite{42-Serban2015HierarchicalNN}  and Distinct metrics (\textbf{Dist-1} and \textbf{Dist-2} ) \cite{43-li-etal-2016-diversity}. Lower perplexity indicates a higher quality of the generated responses. Higher emotion accuracy indicates more precise contextual emotion perception. Larger distinct shows a greater diversity of responses.

\subsubsection{Human Evaluations}

For human evaluation, we employ A/B testing between model response pairs \citep{8-EMPDG,9-KEMP,10-CEM} concerning Empathy (\textbf{Emp.}), Relevance (\textbf{Rel.}), and Fluency (\textbf{Flu.}). Empathy measures the emotional alignment between responses and contexts. Relevance assesses the coherence of generated responses with contexts. Fluency assesses readability and grammar. Three professional annotators compare responses from the CTSM against those from the baselines. Rather than using absolute 1-5 scales, which can be prone to subjective differences \citep{10-CEM}, annotators label CTSM responses as Win, Tie, or Lose relative to the baseline for the same context.

\section{Results and Analysis}

\subsection{Automatic Evaluation Results}

Table \ref{Automatic Evaluation Results} shows the performance of CTSM and baselines concerning automatic metrics. We find that models such as MoEL and MIME focus primarily on recognizing state emotions and exploring the relationship between context and various emotion categories. However, they tend to overlook trait emotions. This oversight decreases emotion detection accuracy (Acc) and compromises response quality (PPL). On the other hand, models like EmpDG, KEMP, CEM and CASE, which only focus on trait emotions, detect emotions more precisely but do not substantially improve response diversity (Dist-1 and Dist-2) or quality (PPL).

Overall, CTSM outperforms all other baselines concerning all the automatic evaluation metrics. Specifically, CTSM achieves a \textbf{7.99\%} relatively higher accuracy than CASE. We attribute this to our model's exceptional capability to combine trait and state emotions, perceiving a more comprehensive range of emotions. Subsequently, through the emotion guidance module, the teacher guides the student to learn features from soft labels that cover both trait and state emotions. This process enhances the emotion encoders' capacity and generates deeper emotional representations.
Furthermore, CTSM achieves a relative improvement over CASE in Dist-1, Dist-2, and PPL by \textbf{170.27\%}, \textbf{83.04\%}, and \textbf{2.29\%} respectively. We attribute the performance improvement to the enhanced integration of the student context and inference knowledge through the emotion guidance module, which enriches response diversity. Furthermore, cross-contrastive learning reduces the distance among generated responses, target responses, and contextual representations regarding both trait and state emotions. This module significantly strengthens the representation of contextual semantics and emotions, enhancing the model's response quality and capacity for empathetic expression.
\begin{table}[!t]
    \centering
    \small
    \begin{tabularx}{\columnwidth}{p{13mm}cccc}
        \toprule
        \textbf{Models} & \textbf{Acc(\%)} $\uparrow$ & \textbf{PPL} $\downarrow$ & \textbf{Dist-1} $\uparrow$ & \textbf{Dist-2} $\uparrow$  \\
        \midrule
        Transformer  & - & 37.73 & 0.47 & 2.04 \\
        MoEL  & 32.00 & 38.04 & 0.44 & 2.10 \\
        MIME  & 34.24 & 37.09 & 0.47 & 1.91 \\
        EmpDG  & 34.31 & 37.29 & 0.46 & 2.02 \\
        KEMP  & 39.31 & 36.89 & 0.55 & 2.29 \\
        CEM  & 39.11 & 36.11 & 0.66 & 2.99 \\
        CASE  & 40.20 & 35.37 & 0.74 & 4.01\\
        \midrule
        CTSM & \textbf{43.41} & \textbf{34.56} & \textbf{2.00} & \textbf{7.34} \\ 
        \bottomrule
    \end{tabularx}
    \caption{Comparison of CTSM against baseline models on automatic evaluation metrics. The best results are bolded.}
    \label{Automatic Evaluation Results}
\end{table}

\begin{table}[!t]
    \centering
    \small
    \begin{tabularx}{\columnwidth}{p{24mm}cccc}
        \toprule
        \textbf{Comparison} & \textbf{Aspects} & \textbf{Win} & \textbf{Lose}  & \textbf{$\kappa$}  \\
        \midrule
        \multirow{3}{*}{CTSM vs. KEMP} & Emp. &\textbf{48.6}& 8.1 & 0.64 \\
                      & Rel. &\textbf{52.6}& 13.3 & 0.60 \\
                      & Flu. &\textbf{35.1}& 14.1 & 0.56 \\
        \midrule
        \multirow{3}{*}{CTSM vs. CEM} & Emp. &\textbf{38.5}& 10.9 & 0.62 \\
                     & Rel. &\textbf{47.9}& 17.0 & 0.68 \\
                     & Flu. &\textbf{28.9}& 15.8 & 0.44 \\
         \midrule
         \multirow{3}{*}{CTSM vs. CASE} & Emp. & \textbf{36.8}& 14.6 & 0.57 \\
           & Rel. & \textbf{49.1} & 16.5 & 0.53 \\
           & Flu. & \textbf{28.1} & 13.8 & 0.48 \\
        \bottomrule
    \end{tabularx}
    \caption{CTSM's human A/B evaluation results(\%). The best results are bolded. $\kappa$ is the label consistency measured by Fleiss' kappa \cite{44-doi:10.1177/001316447303300309}, with 0.41 $\leq$ $\kappa$ $\leq$ 0.60 and 0.61 $\leq$ $\kappa$ $\leq$ 0.80 indicating moderate and substantial agreement respectively.}
    \label{Human Evaluation Results}
\end{table}

\subsection{Human Evaluation Results}

Based on the results shown in Table \ref{Automatic Evaluation Results}, we select three competitive models as benchmarks for human evaluation. As shown in Table \ref{Human Evaluation Results}, CTSM achieves state-of-the-art performance on the human evaluation metrics of Empathy, Relevance, and Fluency compared to others. The high scores in Empathy underscore CTSM's proficiency in perceiving both trait and state emotions within contextual dialogues and in expressing appropriate emotions in the generated responses. Additionally, high scores in Relevance attest to CTSM's capability to comprehend contextual semantics effectively, generate topically coherent responses, and extract and convey pertinent information from diverse contexts. Finally, the outstanding Fluency highlights CTSM's superior decoding capabilities to generate more natural, human-like responses.

\subsection{Ablation Studies}

\begin{table}[!t]
    \centering
    \small
    \begin{tabularx}{\columnwidth}{lp{8.5mm}ccc}
        \toprule
         \textbf{Models} & \textbf{Acc(\%)}  & \textbf{PPL}  & \textbf{Dist-1}  & \textbf{Dist-2}   \\
        \midrule
        CTSM & \textbf{43.41} & 34.56 & \textbf{2.00} & \textbf{7.34} \\
        \midrule
        w/o TEE & 42.58 & 34.68 & 1.18 & 4.28 \\  
        w/o SEE & 42.63 & 35.01 & 1.42 & 5.06 \\ 
        w/o EGM & 39.66 & 35.08 & 1.72 & 6.50  \\ 
        w/o CCL & 42.97 & \textbf{34.42} & 1.73 & 6.27 \\
        w/o EGM \& CCL & 41.88 & 36.35 & 1.57 & 5.42 \\
        \bottomrule
    \end{tabularx}
    \caption{Ablation studies results of CTSM. The best results are bolded.}
    \label{Ablation Studies}
\end{table}

We design four variants for ablation studies to verify the effectiveness of the key components in our model:
\textbf{1) w/o TEE}: Without the \textbf{t}rait \textbf{e}motion embedding and corresponding emotion \textbf{e}ncoder.
\textbf{2) w/o SEE}: Without the \textbf{s}tate \textbf{e}motion embedding and corresponding emotion \textbf{e}ncoder.
\textbf{3) w/o EGM}: Without the \textbf{e}motional \textbf{g}uidance \textbf{m}odule, which contains teacher and student emotion guidance.
\textbf{4) w/o CCL}: Without the \textbf{c}ross-\textbf{c}ontrastive \textbf{l}earning component in the decoding and generating process, and removing the contrastive loss. 
\textbf{5) w/o EGM \& CCL}: Simultaneously eliminating the aforementioned EGM and CCL components.

The results are shown in Table \ref{Ablation Studies}. Omitting \textbf{TEE} from CTSM leads to a notable decline in Acc and Dist, indicating TEE enhances emotion perception accuracy and quality of response.
On the other hand, the exclusion of \textbf{SEE} results in lower emotion accuracy and response quality, highlighting SEE's role in improving alignment between contextual semantics and conveyed emotions, thereby enhancing the model's comprehensive emotion perception and semantic understanding. 
Moreover, the superior Acc in the w/o \textbf{EGM \& CCL} relative to CASE suggests that capturing trait and state emotions alone enhances emotional perception capabilities.

Furthermore, detaching \textbf{EGM} from CTSM significantly reduces all metrics, especially Acc and PPL. It emphasizes that EGM markedly enhances emotion encoders' efficiency and the inference-enriched context's semantic representation.

Finally, CTSM without \textbf{CCL} achieves worse accuracy and diversity. The results demonstrate that CCL enhances the model's ability to fully perceive comprehensive contextual emotions and produce diverse responses, ultimately optimizing empathetic expression. However, its PPL value is better than CTSM. We attribute it to two factors: Firstly, the negative samples contain noise introduced by the dataset, which could be amplified during the feature alignment \citep{35-10.1145/3488560.3498514}. Secondly, a small batch size and relatively short dialogue sentences in the dataset may cause the model to overfit during training \citep{CCL-overfitting-wang2022rethinking}, generating overly simplistic responses and subsequently deteriorating the PPL.

\subsection{Deeper Analysis on Trait and State Emotions}
In this subsection, we delve deeper into the emotional polarities of trait and state, emphasizing the importance of considering them concurrently by showing the discrepancy between them.

Specifically, the word's \textit{trait} emotion polarity $\mathcal{P}_{t}$ is determined by the Valence in the VAD lexicon. Specifically, a word exhibits a negative trait polarity $\mathcal{P}_t=0$ when $0\leq \textnormal{Valence} \leq0.5$, and a positive polarity $\mathcal{P}_t=1$ when $0.5 < \textnormal{Valence} \leq 1$.
Then, we divide all words into positive and negative groups based on their $\mathcal{P}_t$.
For the \textit{state} emotion polarity $\mathcal{P}_s$, we utilize the GloVe word embeddings, which encapsulate rich contextual semantics. 
We calculate the centroid of each of the two groups and the cosine similarity (as in Sec.  \ref{state emotion encoding}) between each word and the two centroids. A word is deemed to have a negative state polarity $\mathcal{P}_s=0$ when it exhibits a more substantial similarity to the negative centroid than to the positive one, and vice versa.

\begin{figure}[!t]
\begin{center}
\includegraphics[width=\columnwidth]{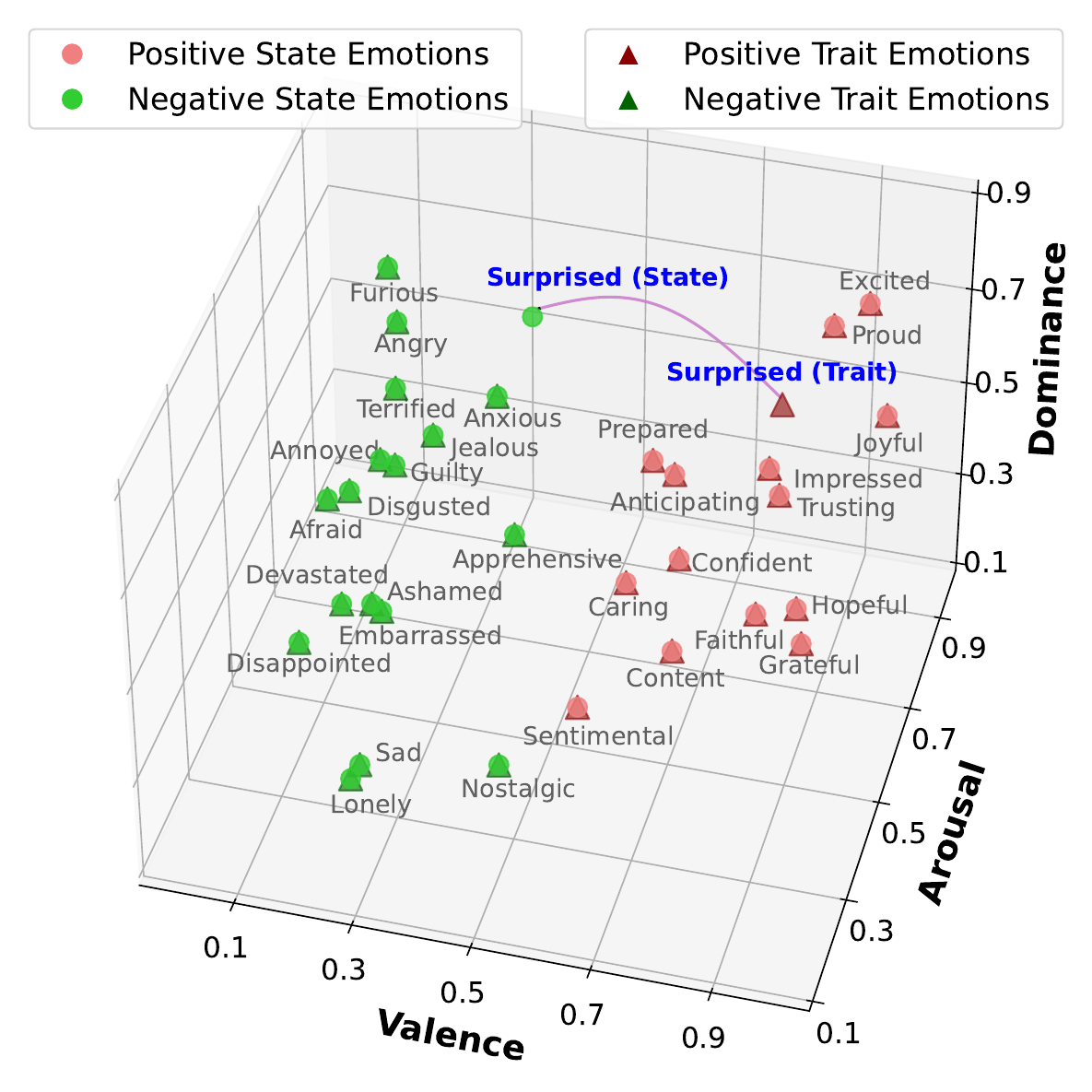} 
\caption{Visualization of trait and state emotion polarities in the VAD  space. When a word's trait and state emotional polarities align, they overlap; otherwise, an offset occurs (highlighted in blue).}
\label{fig.3}
\end{center}
\end{figure}

We visualize the trait and state emotion polarities of 32 emotion words in the VAD three-dimensional space, as illustrated in Figure \ref{fig.3}. Notably, while \textit{surprised} manifests positive trait emotions, its state emotions inclination towards the negative in contextual semantics (whose $\mathcal{P}_t \neq  \mathcal{P}_s$), results in emotion and semantic divergence. Further analysis of 23,712 words from the ED dataset reveals a 48.75\% proportion (11,559 words) exhibiting such discrepancies. Therefore, combining the word's trait and state emotions and analyzing their interplay is necessary to diminish the model's emotion and semantic discrepancies.

\begin{table}[!t]
    \centering
    \footnotesize
    \renewcommand{\arraystretch}{0.9}
    \begin{tabularx}{\columnwidth}{l|X}
        \toprule
        \textbf{Emotion} & Nostalgic \\
        \midrule
        \multirow{3}{*}{\textbf{Context}}& Last week when Toys"R"Us closed, it really made me start thinking of the 90s and my childhood. \\
        \midrule
        \textbf{KEMP} & I am sure you will do great again. \\
        \textbf{CEM}& I bet you were happy for you. \\
        \textbf{CASE} & What kind of memories?\\
        \midrule
        \multirow{3}{*}{\textbf{CTSM}} & That is awesome. I am sure it was a good thing to have a good time to go back to the 90s and the best of us. \\
        \midrule
        \multirow{2}{*}{\textbf{Golden}} & I have heard a lot of people say something similar. Did you go there a lot?  \\
        \bottomrule
    \end{tabularx}
    
    \vspace{0.1em}  

    \centering
    \begin{tabularx}{\columnwidth}{l|X}
        \toprule
        \textbf{Emotion}& Impressed \\
        \midrule
        \multirow{4}{*}{\textbf{Context}} & My 6 year old tried to play a video game but could not understand how the controller worked. A few days later she was playing no problem! \\
        \midrule
        \textbf{KEMP}& I am sure you will do great! \\
        \textbf{CEM}& That is so sweet of her! \\
       \textbf{CASE}& Oh no! Did she have a good time?  \\
        \midrule
        \multirow{2}{*}{\textbf{CTSM}} & That is awesome! I love video games too! \\
        \midrule
        \multirow{3}{*}{\textbf{Golden}} & Kids pick those things up quickly. And it'll help with her hand-eye coordination, reading - all sorts of things! \\
        \bottomrule
    \end{tabularx}

    \caption{Comparison of responses generated by CTSM and three baselines.}
    \label{Case Study}
\end{table}

\subsection{Case Study}
In Table \ref{Case Study}, two case analyses compare CTSM against three prominent baselines: KEMP, CEM and CASE. In the first sample, these baselines fail to integrate the semantics of words such as \textit{closed} and \textit{childhood} with both trait and state emotions, resulting in irrelevant and incoherent responses. In contrast, CTSM effectively merges \textit{nostalgic} trait emotions and state emotions with semantics, inferring that the speaker is reminiscing about the \textit{1990s era}.
In the second sample, the baselines either convey inaccurate emotions (as with CASE) or express weak feelings (as seen with KEMP and CEM). These shortcomings are coupled with a limited grasp of semantics, resulting in generic responses. Given words like \textit{video game} and \textit{not understand} carry negative trait emotions, but \textit{tried} and \textit{no problem} express positive state emotions, CTSM can combine emotions with semantics to understand the feelings conveyed by \textit{impressed}.

\section{Conclusions and Future Work}

In this paper, we propose CTSM that combines trait and state emotions for comprehensive dialogue emotion perception. By encoding trait and state emotion embeddings, CTSM captures dialogue emotions fully. Then, the emotion guidance module further augments emotion perception capability. Lastly, the cross-contrastive learning module enhances the model's empathetic expression capability. The automatic and human evaluation results validate the efficacy of CTSM on the empathetic response generation task. In the future, we will emphasize refining the trait emotion embedding and exploring more methods for state inclination.

\section{Limitations}
Our work primarily has two limitations as follows:

Firstly, we employ a cross-contrastive learning module in CTSM. This process constructs multiple positive sample pairs by cross combining features for contrastive learning. However, it might overlook other strongly correlated positive feature pairs.

The second limitation is the inconsistency between automatic evaluation metrics and human evaluation scores \citep{liu-etal-2016-evaluate}. Automated metrics struggle to assess the degree of empathy in responses, and solely relying on existing metrics makes generating empathetic dialogues challenging.

\section{Ethical Considerations}
The data we use is sourced from E{\small MPATHETIC}\-D{\small IALOGUES} \citeplanguageresource{lr-ED}, an open-source dataset that does not contain any personal privacy information. Our human evaluations are conducted by three professional annotators, ensuring no involvement of personal privacy, with reasonable wages paid. 

\section{Acknowledgements}
We are grateful for the support and assistance from the backers of this project and the related laboratories. 
Yufeng Wang, Zhou Yang, Shuhui Wang, and Xiangwen Liao were supported by the Natural Science Foundation of China (61976054).

\section*{Bibliographical References}\label{sec:reference}

\bibliographystyle{lrec-coling2024-natbib}
\bibliography{CTSM}

\section*{Language Resource References}
\label{lr:ref}
\bibliographystylelanguageresource{lrec-coling2024-natbib}
\bibliographylanguageresource{languageresource}

\end{document}